\documentclass{article}
\usepackage[accepted]{icml2021}
\usepackage[utf8]{inputenc} 
\usepackage[T1]{fontenc}   
\usepackage{hyperref}      
\usepackage{url}            
\usepackage{booktabs}      
\usepackage{amsfonts}     
\usepackage{amsmath}
\usepackage{nicefrac}       
\usepackage{microtype}     
\usepackage{amssymb}       
\usepackage{algorithm} 
\usepackage{algorithmic}  
\usepackage[ruled,vlined,algo2e]{algorithm2e}    
\usepackage{graphicx}  
\usepackage[font={small}]{caption}
\usepackage{subcaption}
\usepackage{todonotes}

\icmltitlerunning{Learning Task Agnostic Skills with Data-Driven Guidance}

\begin{document}

\twocolumn[
\icmltitle{Learning Task Agnostic Skills with Data-driven Guidance}
\icmlsetsymbol{equal}{*}

\begin{icmlauthorlist}
\icmlauthor{Even Klemsdal}{equal,IDI}
\icmlauthor{Sverre Herland}{equal,IDI}
\icmlauthor{Abdulmajid Murad}{equal,IKI}
\end{icmlauthorlist}
\icmlaffiliation{IDI}{Department of Computer Science, Norwegian University of Science and Technology}
\icmlaffiliation{IKI}{Department of Information Security and Communication Technology, Norwegian University of Science and Technology}
\icmlcorrespondingauthor{Even Klemsdal}{even.klemsdal@ntnu.no}

\icmlkeywords{Reinforcement Learning, Unsupervised Reinforcement Learning, Skill discovery}
\vskip 0.3in
]

\printAffiliationsAndNotice{\icmlEqualContribution} 

%%%%%%%%%%%%%%%%%%%%%%%%%%%%%%%%%%%%%%%%%%%%%%%%%%%%%%%%%%%%%%%%%%%%%%%%%%%%%%%%%%%%%%%5%%
\begin{abstract}
To increase autonomy in reinforcement learning, agents need to learn useful behaviours without reliance on manually designed reward functions. To that end, skill discovery methods have been used to learn the intrinsic options available to an agent using task-agnostic objectives. However, without the guidance of task-specific rewards, emergent behaviours are generally useless due to the under-constrained problem of skill discovery in complex and high-dimensional spaces. This paper proposes a framework for guiding the skill discovery towards the subset of expert-visited states using a learned state projection. We apply our method in various reinforcement learning (RL) tasks and show that such a projection results in more useful behaviours.
\end{abstract}

%%%%%%%%%%%%%%%%%%%%%%%%%%%%%%%%%%%%%%%%%%%%%%%%%%%%%%%%%%%%%%%%%%%%%%%%%%%%%%%%%%%%%%%
\section{Introduction}
 \label{sec:introduction}
While autonomous learning of diverse and complex behaviors is challenging, significant progress has been made using deep reinforcement learning (DRL). The progress has been accelerated by the powerful representational learning of deep neural networks ~\cite{lecun2015deep}, and the scalability and efficiency of RL algorithms ~\cite{mnih2015human, schulman2017proximal, haarnoja2018soft, lillicrap_continuous_2019}.
However, DRL still involves an externally designed reward function that guides learning and exploration. Manually engineering such a reward function is a complex task that requires significant domain knowledge in such a way that hinders autonomy and adoption of RL. 
Prior works have proposed using unsupervised skill discovery to alleviate these challenges by using empowerment as an intrinsic motivation to explore and acquire abilities \cite{salge2014empowerment, gregor2016variational, eysenbach2018diversity, sharma2019dynamics,campos2020explore}.

Although skill discovery without a reward function can be helpful as a primitive for downstream tasks, most of the emergent behaviours in the learned skills are useless or of little interest. This is a direct consequence of \textit{under-constrained} skill discovery in complex and high dimensional state space. One possible solution is to leverage prior knowledge to bias skill discovery towards a subset of the state space through a hand-crafted transformation of the state space \cite{eysenbach2018diversity, campos2020explore}. However, utilizing prior knowledge to hand-craft such a transformation contradicts the primary goal of unsupervised RL, which is reducing manual design efforts and reliance on prior knowledge.

Instead, we explore how to learn a parameterized state projection that directs skill discovery towards the subset of expert-visited states. To that end, we employ examples of expert data to train a state encoder through an auxiliary classifier, which tries to distinguish expert-visited states from random states. We then use the encoder to project the state space into a latent embedding that preserves information that makes expert-visited states recognizable. This method extends readily to other mechanisms of learned state-projections and different skill discovery algorithms. Crucially, our method requires only samples of expert-visited states, which can easily be obtained from any reference policy, for example expert demonstrations.

The key contribution of this paper is a simple method for learning a parameterized state projection that guides skill discovery towards a substructure of the observation space. We demonstrate the flexibility of our state-projection method and how it can be used with the skill-discovery objective. We also present empirical results that show the performance of our method in various locomotion tasks.

%%%%%%%%%%%%%%%%%%%%%%%%%%%%%%%%%%%%%%%%%%%%%%%%%%%%%%%%%%%%%%%%%%%%%%%%%%%%%%%%%%%%%%%
\section{Related Work}
\label{sec:related-work}

Unsupervised reinforcement learning aims at learning diverse behaviours in a task-agnostic fashion without guidance from an extrinsic reward function~\cite{jaderberg2016reinforcement}. This can be accomplished through learning with an intrinsic reward such as \textit{curiosity} \cite{oudeyer2009intrinsic} or \textit{empowerment}  ~\cite{salge2014empowerment}. The notion of curiosity has been utilized for exploration by using predictive models of the observation space and providing a higher intrinsic reward for visiting unexplored trajectories ~\cite{pathak2017curiosity}. Empowerment addresses maximizing an agent's control over the environment by exploring states with maximal intrinsic options (skills). 

Several approaches have been proposed in the literature to utilize empowerment for skill discovery in unsupervised RL. Gregor et al. \cite{gregor2016variational} developed an algorithm that learns intrinsic skill embedding and used generalization to discover new goals. They used the mutual information between skills and final states as the training objective and hence used a discriminator to distinguish between different skills. Eysenbach et al. \cite{eysenbach2018diversity} used mutual information between skills and states as an objective while using a fixed embedding distribution of skills. Additionally, they used a maximum-entropy policy \cite{haarnoja2018soft} to produce stochastic skills. However, most of the previous approaches assume a state distribution induced by the policy itself, resulting in a premature commitment to already discovered skills. Campos et al. \cite{campos2020explore} used a fixed uniform distribution over states to break the dependency between the state distribution and the policy.

Certain prior work has addressed the challenge of complex and high dimensional state space by constraining the skill-discovery in a subset of the state space. Sharma et al.  \cite{sharma2019dynamics} learned predictable skills by training a skill-conditioned dynamic model instead of a discriminator to model specific behaviour in a subset of the state space. Eysenbach et al. \cite{eysenbach2018diversity} proposed incorporating prior knowledge by conditioning the discriminator on a subset of the state space using a hand-crafted and a task-specific transformation. Our work addresses this challenge by guiding the skill discovery towards the subset of expert-visited states. In contrast to inverse reinforcement learning, \cite{fu2018learning}, we do not explicitly infer the extrinsic reward.  Crucially, we do not try to learn the expert policy directly in contrast to behaviour cloning or imitation learning \cite{ross2011reduction}. Our proposed method resembles the algorithm proposed by Li et al. \cite{li2020reinforcement} in which they used a Bayesian classifier that estimates the probability of successful outcome states, resulting in a more task-directed exploration. However, their algorithm does not optimize the mutual information; hence it does not learn diverse skills via the discriminability objective.

%%%%%%%%%%%%%%%%%%%%%%%%%%%%%%%%%%%%%%%%%%%%%%%%%%%%%%%%%%%%%%%%%%%%%%%%%%%%%%%%%%%%%%%
\section{Preliminaries}
\label{sec:preliminaries}

In this paper, we formalize the problem of skill discovery as a Markov decision process (MDP) without a reward function: $\mathcal{M}=(\mathcal{S},\mathcal{A},\mathcal{P})$, where $\mathcal{S}$ is the state space, $\mathcal{A}$ is the action space, and  $\mathcal{P}: \mathcal{S} \times \mathcal{S} \times \mathcal{A} \rightarrow [0, \infty)$  is the transition probability density function. The RL agent learns a skill-conditioned policy $\pi_{\vartheta}(a | s, z)$, where the skill $z$ is sampled from some distribution $p(z)$. A skill, or option (as first introduced in \cite{sutton2018reinforcement}), is a temporal abstraction of a course of actions that extends over many time steps. We will also consider the information-theoretic notion of mutual information between states and skills $\mathcal{I}(S; Z) = \mathcal{H}(Z) - \mathcal{H}(Z | S) $, where $\mathcal{H}(\cdot)$ is the Shannon entropy.

\subsection{Skill Discovery Objective}

The overall goal of skill discovery is to find a policy $\pi_\vartheta$ capable of carrying out different tasks that are learned without extrinsic supervision for each type of behavior. 
We consider policies of the form $\pi_\vartheta(a|s,z)$ that specify different distributions over actions depending on which skill $z$ they are conditioned on.
Although this general framework does not constrain how $z$ should be represented, we define it as a discrete variable since it has been empirically shown to perform better than continuous alternatives \cite{eysenbach2018diversity}. 

We follow the framework proposed by the "diversity is all you need" (DIAYN) algorithm \cite{eysenbach2018diversity}, in which skills are learned by defining an intrinsic reward that promotes diversity.
Intuitively, each skill should make the agent visit a unique section of the state space.
This can be expressed as maximising the mutual information $\mathcal{I}(S;Z)$ of the state visitation distributions for different skills \cite{salge2014empowerment}.
To ensure that the visited areas of the state space are spaced sufficiently far apart, we use a soft policy that maximises the entropy of the action distribution.
Formally, we maximize the following objective function:
 
\begin{equation}
    \begin{split}
        \mathcal{F}(\theta) &\triangleq I(S;Z) + \mathcal{H}(A \vert S) - I(A; Z \vert S) \\
        &= \mathcal{H}[A \vert S,Z] - \mathcal{H}[Z \vert S] + \mathcal{H}[Z] \\
    \end{split}
    \label{eq:objective}
\end{equation}

The first term $\mathcal{H}[A \vert S,Z]$ means that the policy should act as randomly as possible and can be optimized by maximizing the policy's entropy.
The second term $-\mathcal{H}[Z \vert S]$ dictates that each visited state should (ideally) identify the current skill. 
The third term is the entropy of the skill distribution, which can be maximized by deliberately sampling skills from a uniform distribution during training.
Unfortunately, $-\mathcal{H}[Z \vert S]$ requires knowledge about $p(z \vert s)$, which is not readily available.
Consequently, we approximate the true distribution by training a classifier $q_{\phi}(z \vert s)$, leading to a lower bound:

\begin{equation}
    \begin{split}
        \mathcal{F}(\theta) &= \mathcal{H}[A \vert S,Z] + \mathbb{E}_{p}[\log p(z \vert s)] - \mathbb{E}_{p}[\log p(z)] \\
        &\geq \mathcal{H}[A \vert S,Z] + \mathbb{E}_{p}[\log q_{\phi}(z \vert s)] - \mathbb{E}_{p}[\log p(z)] 
    \end{split}
    \label{eq:objective-lb}
\end{equation}
The lower bound follows from the non-negative property of the Kullback-Leibler divergence $D_{KL}(p \vert \vert q) = \mathbb{E}_{p}[\log p(z|s) - \log q(z|s)] \ge 0$, which can be rearranged to $\mathbb{E}_p[\log p(z|s)] \ge \mathbb{E}_p[\log q(z|s)]$ \cite{agakov2004algorithm}. 

The classifier $q_\phi$ is fitted throughout training with maximum likelihood estimation over the sampled states and active skills. This leads to a scenario where the policy is rolled out for a (uniformly) sampled skill, and the classifier is trained to detect the skill based on the states that were visited. The policy is given a reward proportional to how well the classifier could detect the skill in each state.  In the end, this should make the policy favor visiting disjoint sets of states for each skill, leading to a cooperative game between $q_\phi$ and $\pi_\vartheta$.

\subsection{Limitations of existing methods}

\begin{figure}[h!]
  \centering
  \includegraphics[width=0.48\textwidth]{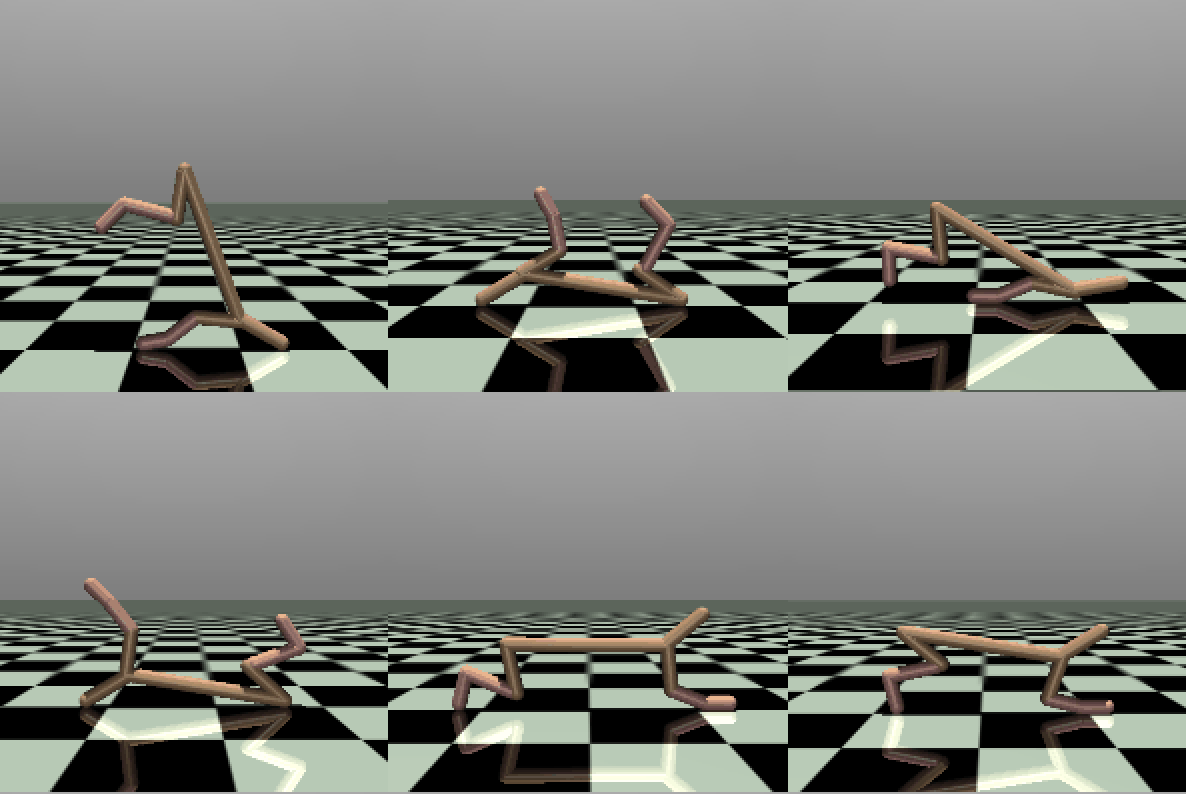}
  \caption{Cheetahs frozen in various configurations for 6 unique skills. When sampled, the policy generates actions that "shake" around these positions. However, they are still distinguishable by the classifier $q_\phi$ with over 90\% accuracy.}
  \label{fig:frozen-cheetas}
\end{figure}

A major challenge that arises when maximizing the objective in Equation \ref{eq:objective-lb}, particularly in applications with high-dimensional spaces, is that it becomes trivial for each skill to find a sub-region of the state space where it is easy to be recognised by $q_\phi$. In preliminary experiments, we observed that the existing methods discovered behaviours that covered small parts of the state space. For the HalfCheetah environment \cite{brockman2016openai} this resulted in many skills generating different types of static poses (see Figure \ref{fig:frozen-cheetas}) and not many skills exhibiting "interesting" behaviour such as locomotion.

Optimising for $\mathcal{H}[A \vert S, Z]$ should mitigate this issue to some extent. Increasing the policy's entropy incentivises the skills to progressively visit regions of the state space that are so far apart that not even highly stochastic actions will cause them to overlap accidentally.
However, it has been shown that mutual information based algorithms have difficulties spreading out to novel states due to low values of $\log q_\phi(z|s)$ for out-of-sample states \cite{campos2020explore}. 

%%%%%%%%%%%%%%%%%%%%%%%%%%%%%%%%%%%%%%%%%%%%%%%%%%%%%%%%%%%%%%%%%%%%%%%%%%%%%%%%%%%%%%%
\section{Proposed Method}
\label{sec:method}

The main idea of our approach is to focus the skill discovery towards certain parts of the state space by using expert data as a prior.
The DIAYN algorithm can be biased towards a user-specified part of the state space by changing the discriminator to maximize $\mathbb{E}[\log q_\phi(z \vert f(s))]$, where $f$ represents some transformation of the state space \cite{eysenbach2018diversity}. Instead of using a hand-crafted $f$ to improve the skills discovered for a navigation task, we aim to learn a parameterized $f_\chi$ by using expert data.

\subsection{State Space Projections}

We consider linear projections of continuous factored state representations on the form $f_\chi : \mathbb{R}^{|S|} \xrightarrow[]{} \mathbb{R}^{|E|}$ with $e = f_\chi(s) = \chi s$, $\chi \in \mathbb{R}^{|E| \times |S|}$, and $|E| < |S|$.
In principle, the idea should apply to more complex mappings, such as a multi-layer perceptron. However, we want to limit the scope of skill discovery to a hyperplane within the original state space.

For the same reason, we also omit any non-linearities in the encoder.
Squeezing the output through a Sigmoidal function would limit discriminability at the (potentially interesting) extremes of the encoding.
Similarly, a ReLU function would effectively eliminate all exploration along the negative direction of $f_{\chi_i}$.
In summary, the objective of the DIAYN skill classifier becomes:
\begin{equation}
    \max_\phi \mathbb{E}[\log q_\phi(z \vert e)]
    \label{eq:clf-objective-enc}
\end{equation}
We learn the parameters $\chi$ for the projection through an auxiliary discriminative objective. 
Specifically, a binary classifier $h_\psi : \mathbb{R^{|E|}} \xrightarrow[]{} \{0, 1\}$ is trained to predict whether an (encoded) state was sampled from the marginal state visitation distribution of a random policy $\pi_{rand}$ or from the distribution of a reference (expert) policy $\pi^*$.
Let $x \sim \mathcal{D}$ denote whether a state $s$ was visited by the reference policy or not (in dataset $\mathcal{D}$), then the parameters of $f_\chi$ are obtained through joint pretraining with $h_\psi$ by maximizing the log likelihood over $\mathcal{D}$:

\begin{equation}
    \max_{\chi, \psi }  \mathbb{E}_{x,s \sim \mathcal{D}}[x \log h_\psi(x | e) + (1 - x)\log (1 - h_\psi(x | e))]
    \label{eq:disc-pretrain}
\end{equation}

where the dataset $\mathcal{D}$ is collected prior to training the main RL algorithm. 
The first half (random samples) are collected by rolling out $\pi_{rand}$ whereas the second half (reference samples) are collected by rolling out $\pi^*$.
After the objective in Equation \ref{eq:disc-pretrain} is optimized, the discriminator $h_\psi$ is discarded and the projection encoding $f_\chi(s)$ is extracted to be used for the objective in Equation \ref{eq:clf-objective-enc}.
Analogous to autoencoders \cite{hinton_reducing_2006}, the idea is that the embeddings produced by $f_\chi(s)$ should now contain a more compact representation of the state space without collapsing the dimensions that make "interesting" behaviour stand out.

While the use of a reference data changes our approach from a strictly unsupervised skill discovery algorithm, the discriminative objective in equation \ref{eq:disc-pretrain} resembles the objectives used in adversarial inverse reinforcement learning (e.g. \cite{fu2018learning}).
However, it differs in that it makes no attempts at matching the behaviour of a reference policy as it is used only as a prior for simplifying the state space.
This approach could also be used with samples from several different reference policies with substantially different marginal state distributions.
As long as their variation can be explained sufficiently without full use of the entire state space, a projection should simplify skill discovery.

\subsection{Implementation}

For learning diverse skills, we use DIAYN as a basis framework. DIAYN uses the Soft Actor-Critic (SAC) algorithm \cite{haarnoja2018soft} that is optimized using policy gradient style updates in contrast to the reparameterized version (DDPG style updates \cite{lillicrap_continuous_2019}). They also use a Squashed Gaussian Mixture Model to represent the policy $a \sim \pi_\vartheta = \tanh GMM(\mu_\vartheta(s), \sigma_\vartheta(s))$. The learning objective is to maximize the mutual information between the state and skill $I(S; Z)$. This objective is optimized by replacing the task rewards with a pseudo-reward

\begin{equation}
\label{eq:reward}
 r_z(s,a) \triangleq \log p_\phi(z \vert s) - \log p(z)   
\end{equation}

where $q_\phi$ is trained to discriminate between skills and p(z) is the fixed uniform prior over skills \cite{eysenbach2018diversity}. A skill is sampled from $z \sim p(z)$ and used throughout a full episode.

In contrast to DIAYN, we use two Q-functions $Q^1_\theta(s,a)$ \& $Q^2_\theta(s,a)$ where both Q-functions attempt to predict the same quantity. This allows us to sample differentiable actions and climb the gradient of the minimum of the two Q-functions (DDPG-style update \cite{lillicrap_continuous_2019}), giving us this objective:
\begin{align*}
        J(\pi_\vartheta) = \min_{i \in \{1, 2\}} Q^i_{\theta}(\pi_\vartheta(a|s)) + \alpha \mathcal{H}(\pi_\vartheta(a|s))
\end{align*}
Like in DIAYN, we also use a Squashed Gaussian Mixture Model to promote diverse behaviour. 

\begin{figure}[h!]
\centering
  \includegraphics[scale=0.255]{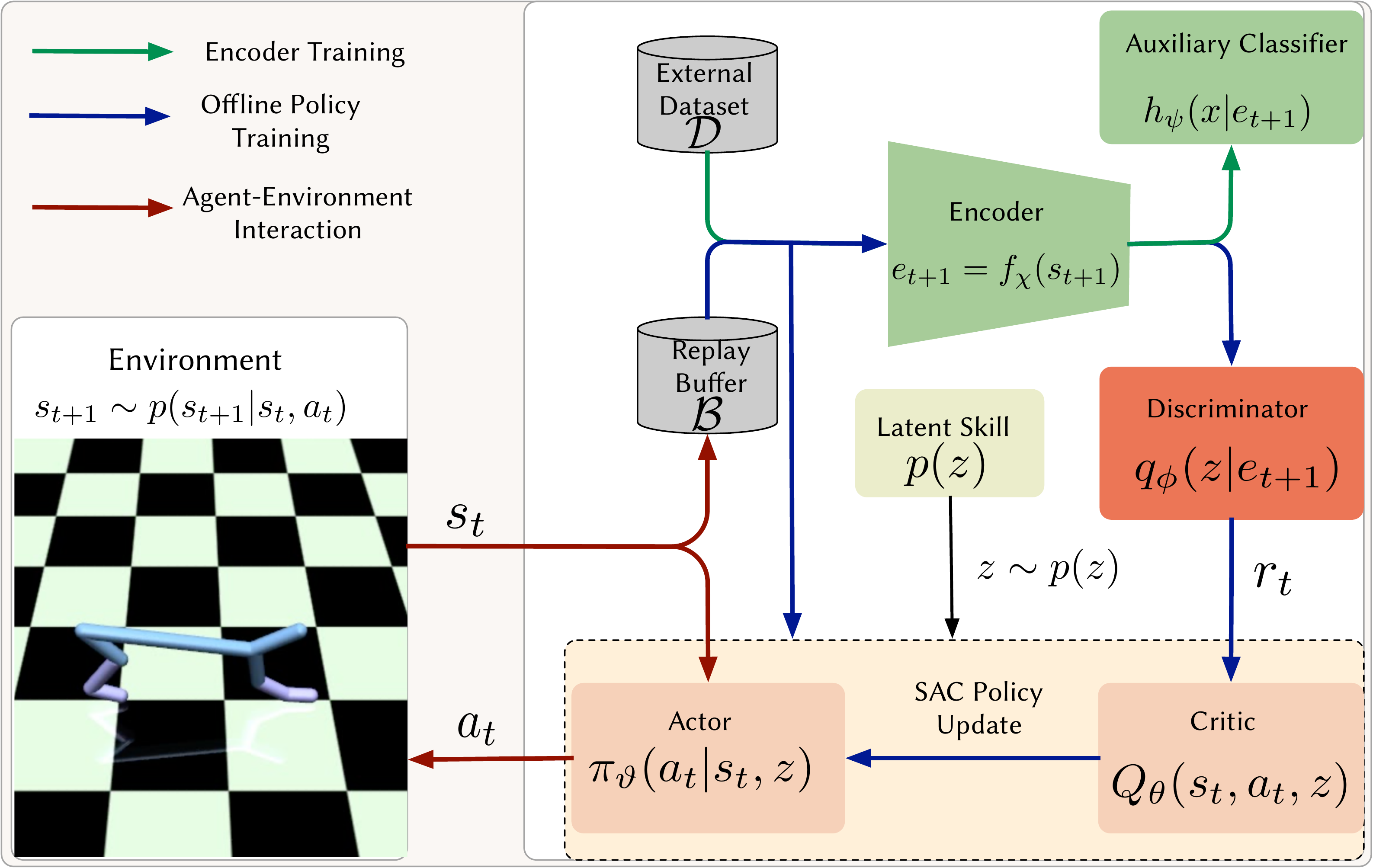}
  \caption{Framework for training expert-guided skill discovery. Green arrows show the training of the encoder, red arrows show the agent-environment interaction, while the blue arrows show the interactions for the offline policy training. Note there is no gradients through the encoder in the offline training.}
  \label{fig:system_architecture}
\end{figure}

\autoref{fig:system_architecture} illustrates the training process of the proposed expert-guided skill discovery. First, we train the encoder $f_\chi$ jointly with the auxiliary classifier $h_\psi$ using the external dataset $\mathcal{D}$. Secondly, we  train the agent using an offline policy algorithm (SAC), in which the agent samples a skill $z \sim p(z)$, and then interacts with the environment by taking action $a_t$ according the skill-conditioned policy $\pi_\vartheta(a_t|s_t, z)$. The environment, then, transits to a new state according to the transition probability $s_{t+1} \sim p(s_{t+1}|s_t, a_t)$. We add this transition $(s_z, z, a_t, s_{t+1})$ to the replay buffer $\mathcal{B}$. Simultaneously, the policy is updated by sampling a mini-batch from the replay buffer $\mathcal{M} \sim \mathcal{B}$, then encoding the next states $e_{t+1} = f_\chi(s_{t+1})$ and passing them through the discriminator $q_\phi(z|e_{t+1}$ to get the intrinsic reward. This reward is used by the Q-functions $Q_\theta(s_t, a_t, z)$ to minimize the soft Bellman residual and update the policy. A pseudocode for the proposed approach can be found in the supplementary material. 

%%%%%%%%%%%%%%%%%%%%%%%%%%%%%%%%%%%%%%%%%%%%%%%%%%%%%%%%%%%%%%%%%%%%%%%%%%%%%%%%%%%%%%%
\section{Experiments}
\label{sec:experiments}

In our experimental evaluation, we aim to demonstrate the impact of our approach of restricting skill discovery to a projection subspace. 
We verify our method on both point-mazes and continuous control locomotion tasks. All the code for running the experiments are publicly available on GitHub\footnote{Project code base: \url{https://github.com/sherilan/cs285-project/tree/master}}.

\subsection{Point Maze}

As an illustrative example, we begin by testing the algorithm on a simple 2D point-maze problem.
The term \textit{maze} is used very generously here, as the environment consists of an open plane in $\mathbb{R}^2$ enclosed by walls that restrict the agent to $\frac{1}{7} < x < \frac{6}{7}$ and $\frac{1}{7} < y < \frac{6}{7}$.
At initialization, the agent is dropped down at $x,y = \frac{7 + \epsilon}{14}, \epsilon \sim U(-1, 1)$ and incentivized to move towards the lower right by a reward proportional to a gaussian kernel centered at ($\frac{9}{14}$, $\frac{3}{14}$).
The agent is free to move by $\pm \frac{1}{70}$ in both the x and y direction.

\begin{figure}[!htp]
    \centering
    \includegraphics[trim=10 20 0 13, width=0.4\textwidth]{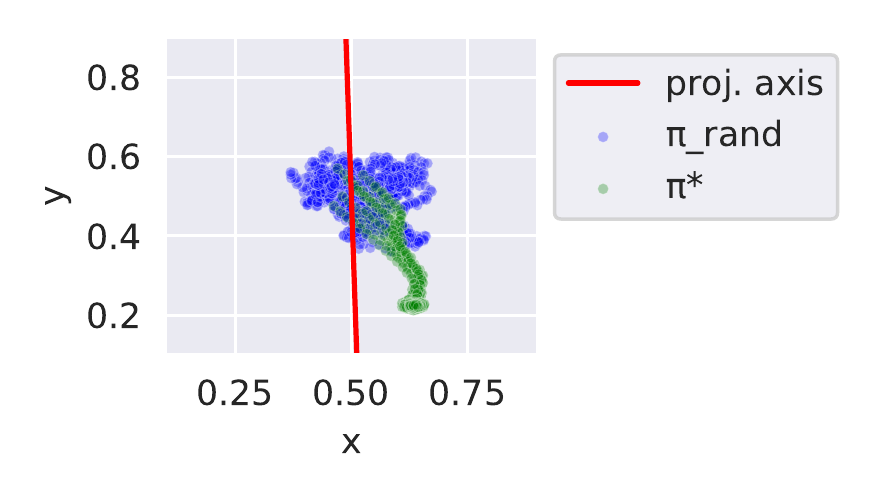}
    \caption{The learned projection axis (in red), which is the output of the encoder when trained from the discrimination of the green and blue points.}
    \label{fig:point-maze-proj}
\end{figure}

\begin{figure}
    \centering
    \includegraphics[trim=10 20 0 10, width=0.5\textwidth]{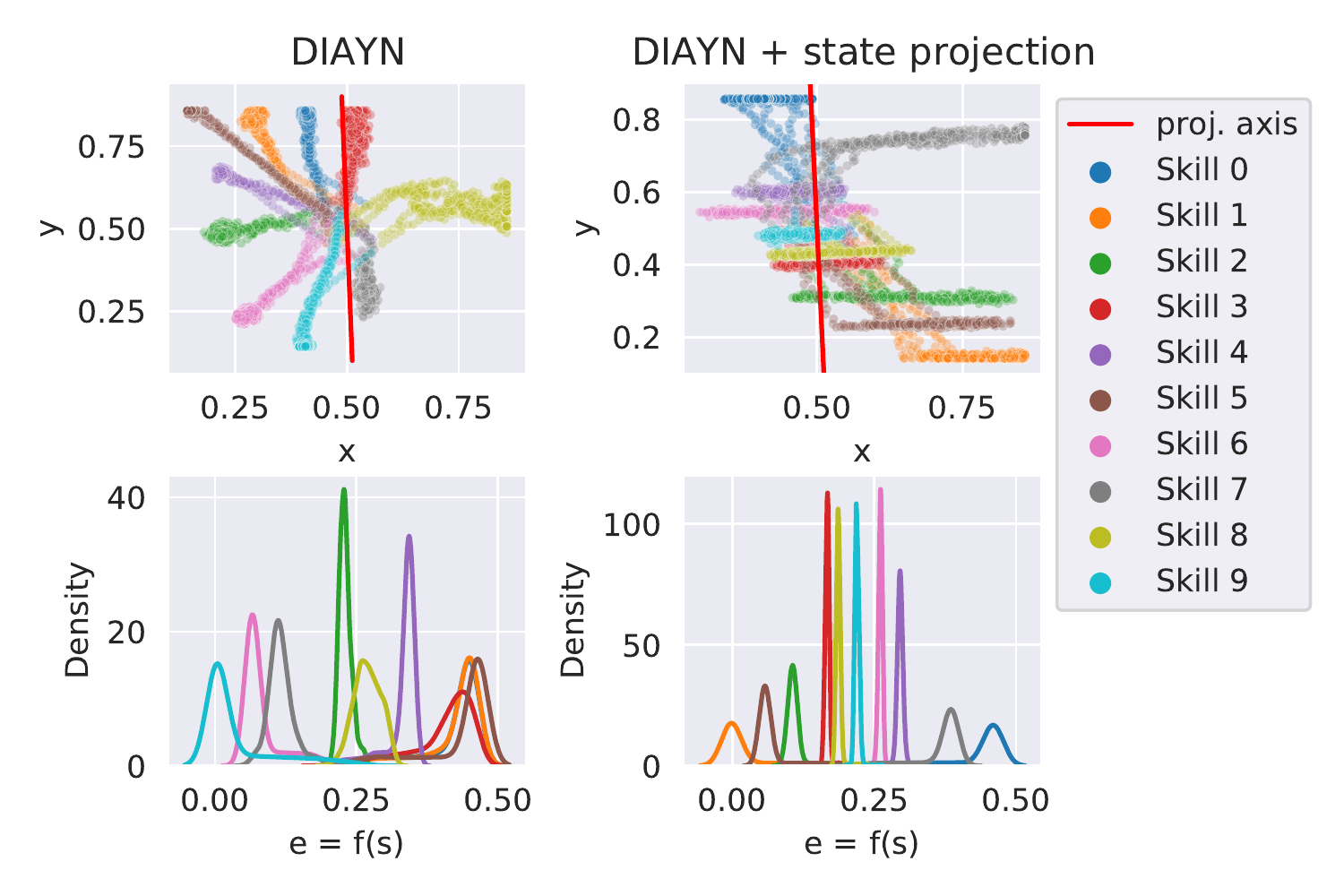}
    \caption{State visitation visualization for 10 skills learned with baseline DIAYN (left) and DIAYN with state projection (right). The top row shows the actual states visited in $\mathbb{R}^2$, whereas the bottom row shows the same data linearly mapped onto the projection axis obtained in Figure \ref{fig:point-maze-proj}. 
    }
    \label{fig:point-maze-diayn}
\end{figure}

We train a SAC agent against the extrinsic environment reward to convergence and set its final policy as the reference policy $\pi*$.
We then sample 10 trajectories of length 100 (green dots in Figure \ref{fig:point-maze-proj}) with $\pi*$, as well as 10 trajectories of length 100 (blue dots in Figure \ref{fig:point-maze-diayn}) from a uniform random policy $\pi_{rand}$.
The resulting dataset $\mathcal{D}$ consists of 2000 samples and is used to train $h_\psi(x | e)$ until it can distinguish states from $\pi*$ and $\pi_{rand}$ with around 98\% accuracy.
For this experiment, we project down from 2D to 1D, making $\chi \in \mathbb{R}^{1 \times 2}$.
The resulting projection axis is visualized as a red line in Figure \ref{fig:point-maze-proj} and is the only thing exported to the next stage of the algorithm.

We then train two versions of the DIAYN algorithm; a baseline using the states as-is in the classifier ($q_\phi(z \vert s)$) and our proposed method using the state projections ($q_\phi(z \vert f_\chi(s))$).
All other hyperparameters are held equal in the two experiments, and the algorithms are trained for 400,000 environment interactions, each attempting to learn 10 distinct skills.

Figure \ref{fig:point-maze-diayn} visualizes the results of the baseline and the state projection to the left and right, respectively.
The top row shows the states that were visited for five rollouts of each skill.
As expected, the baseline skills spread out in all directions (albeit slightly more so towards the left) and converge on locations that are easy to distinguish with a 2D state representation. 
In contrast, the skills generated in the left plot form lines along the state projection axis.
Their wide lateral spread follows from the (unbounded) entropy maximization objective. Besides, any movement perpendicular to the projection axis will not affect the 1D vector passed to the classifier.

\subsection{Mujoco Environments}

Next, we evaluate the algorithm on three continuous control problems from the OpenAI gym suite. \cite{brockman2016openai}. The environments include \textit{HalfCheetah} ($S=\mathbb{R}^{17}$, $A=\mathbb{R}^6$), \textit{Hopper} ($S=\mathbb{R}^{11}$, $A=\mathbb{R}^3$) and \textit{Ant} ($S=\mathbb{R}^{27}$, $A=\mathbb{R}^8$). We choose these environments because they involve substantially different locomotion methods. Additionally, they have observation spaces with different dimensionality, which enable us to better investigate the projection impact.

\begin{figure}[htp]
    \centering
    \includegraphics[trim=10 10 10 0, width=0.4\textwidth]{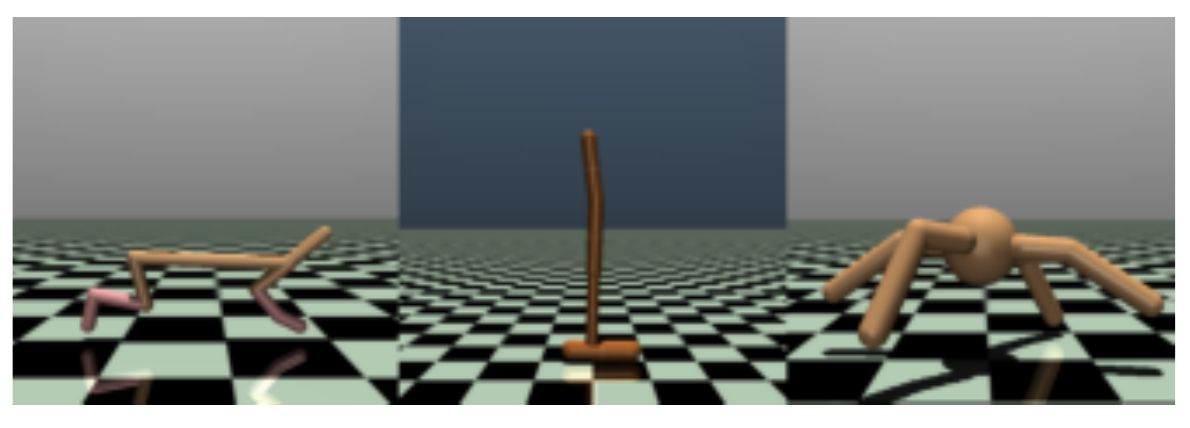}
    \caption{Continuous control environments from the Mujoco part of the OpenAI gym suite. Left: HalfCheetah-v2. center: Hopper-v2, right: Ant-v2.}
    \label{fig:envs}
\end{figure}

\begin{figure*}[!htb]
    \centering
    \includegraphics[trim=0 20 0 10, width=0.9\textwidth]{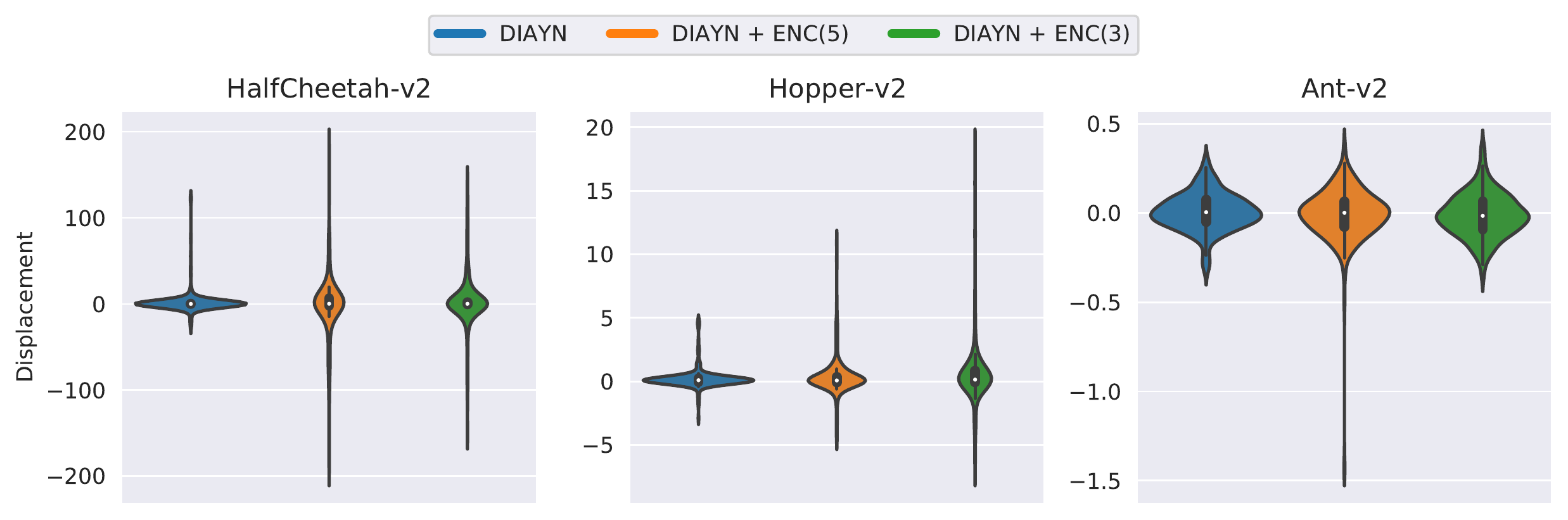}
    \caption{Impact of state-space projection illustrated as distribution of displacements along the locomotion axis used to calculate the extrinsic reward of the environments. 
    By introducing a projection step, the skill search gets focused towards locomotion skills, resulting in a larger spread of displacements.
    }
    \label{fig:pos-dist}
\end{figure*}

\begin{table*}[!htb]
\tiny
\caption{Summary statistics for displacement (along the locomotion axis used to calculate the extrinsic reward of the environments) across the 50 skills learned. The values to the right of $\pm$ indicate standard deviation across 5 seeded runs.}
\resizebox{\textwidth}{!}{%
\begin{tabular}{llccccc}
    \toprule
          &                &               min &             25\% &            50\% &             75\% &              max \\
    \midrule
    HalfCheetah-v2 & DIAYN &    -9.7 ± 11.1 &  -0.1 ± 0.0 &   0.0 ± 0.1 &  0.3 ± 0.1 &   76.6 ± 46.6 \\
          & DIAYN + ENC(3) &   -88.6 ± 42.1 &  -0.4 ± 0.4 &   0.2 ± 0.1 &  3.9 ± 4.7 &   99.0 ± 37.5 \\
          & DIAYN + ENC(5) &  -129.0 ± 48.5 &  -6.1 ± 9.5 &   0.7 ± 1.2 &  6.6 ± 5.1 &  121.2 ± 44.7 \\
    \midrule
    Hopper-v2 & DIAYN &     -1.0 ± 1.1 &  -0.0 ± 0.1 &   0.1 ± 0.0 &  0.2 ± 0.1 &     3.7 ± 1.4 \\
          & DIAYN + ENC(3) &     -4.3 ± 1.8 &  -0.0 ± 0.1 &   0.2 ± 0.2 &  0.9 ± 0.6 &    10.1 ± 6.1 \\
          & DIAYN + ENC(5) &     -3.1 ± 1.4 &  -0.1 ± 0.0 &   0.1 ± 0.0 &  0.4 ± 0.1 &     7.0 ± 3.2 \\
    \midrule
    Ant-v2 & DIAYN &     -0.3 ± 0.0 &  -0.1 ± 0.0 &   0.0 ± 0.0 &  0.1 ± 0.0 &     0.3 ± 0.1 \\
          & DIAYN + ENC(3) &     -0.3 ± 0.1 &  -0.1 ± 0.0 &  -0.0 ± 0.0 &  0.1 ± 0.0 &     0.3 ± 0.1 \\
          & DIAYN + ENC(5) &     -0.5 ± 0.5 &  -0.1 ± 0.0 &  -0.0 ± 0.0 &  0.1 ± 0.0 &     0.3 ± 0.1 \\
    \bottomrule
\end{tabular}}
\label{tab:returns}
\end{table*}
We do one baseline run without any state projection for all three problems, one with a projection down to $\mathbb{R}^5$, and one with a projection down to $\mathbb{R}^3$.
We use our base SAC implementation to obtain reference policies $\pi^*$ and sample 10 trajectories of length 1000 with fairly high returns (Ant: $5063.9 \pm 469.8$, Cheetah: $10656.4 \pm 673.5$, Hopper: $3348.0 \pm 316.0$).
The DIAYN algorithm is otherwise identical to \cite{eysenbach2018diversity} in terms of hyperparameters; $Q$, $\pi$, $q_\phi$, and $h_\psi$ use MLP architectures with 2 hidden layers of width 300, the entropy bonus weight $\alpha$ is set to 0.1, and the number of skills is set to 50.
We limit each skill-discovery run to 2.5 million environment interactions but repeat each experiment 5 times with different random seeds (including training of SAC agents for $\pi^*$).

For quantitative evaluation, we look at the displacement along the target locomotion axis for the extrinsic objective. 
In our approach, we would expect to observe skills that cover this axis well, i.e., skills that run forward and backward at different speeds.
To test this, we roll out each skill deterministically\footnote{Deterministic sampling from our GMM-based policy implies taking the mean of the component with the highest mixture probability.}, record its movement over 1000 time steps (or until it reaches a terminal state) and observe the inter-skill spread.
A similar assessment is possible by only looking at the environment's rewards. 
However, the environment reward also includes terms for energy expenditure, staying alive (for Ant/Hopper), and collisions (Ant), which would obscure the results. 
Figure \ref{fig:pos-dist} shows the displacement distribution of the 50 skills across all runs. 
The same information is summarized numerically in Table \ref{tab:returns}. 

For a qualitative evaluation, we have also composed a video with every skill across all runs\footnote{Video of skills: \url{https://www.youtube.com/watch?v=Xx7RVNmv1tY}}.

%%%%%%%%%%%%%%%%%%%%%%%%%%%%%%%%%%%%%%%%%%%%%%%%%%%%%%%%%%%%%%%%%%%%%%%%%%%%%%%%%%%%%%%
\section{Discussion}

For HalfCheetah and Hopper, the runs with state encoding (+ ENC(3|5)) exhibit a substantially larger spread than the baseline.
The best forward-moving cheetah skill moves 178 units forward ($=3311$ environment return), and the best backwards-moving cheetah skill moves 186 units backwards ($=-4025$ environment return).
For the hopper environment, the best forward-moving skill manages to jump 20 units forward, which corresponds to an environment reward of 3268, which is on the same level as the reference data used to fit its encoder.

The results in the Ant environment are less impressive. 
There is hardly any difference in how the displacements are distributed for the three approaches, and the total movement is almost negligible.
For reference, a good Ant agent trained against the extrinsic reward should obtain displacements in the 100s when evaluated over the same trajectory horizon.

Looking at the generated Ant behaviour, we found that the skills produced with encoders typically moved even less than those generated by the baseline.
This is not because it is impossible to generate a linear projection that promotes locomotion at various speeds, as the state representation of all three problems contains a feature for linear velocity along the target direction.
Moreover, the skill classifier does reach a high accuracy (some breaking 90\%), so the algorithm manages to find distinguishable skills.
We, therefore, suspect that the procedure used to fit the encoder is insufficient for this environment. 
While it does pick up on linear velocity, it also picks up on several other features from the state space, which might have made it easier for the algorithm to make the skills distinguishable. 

\begin{figure}[h!]
    \centering
    \includegraphics[trim=10 10 10 10, width=0.48\textwidth]{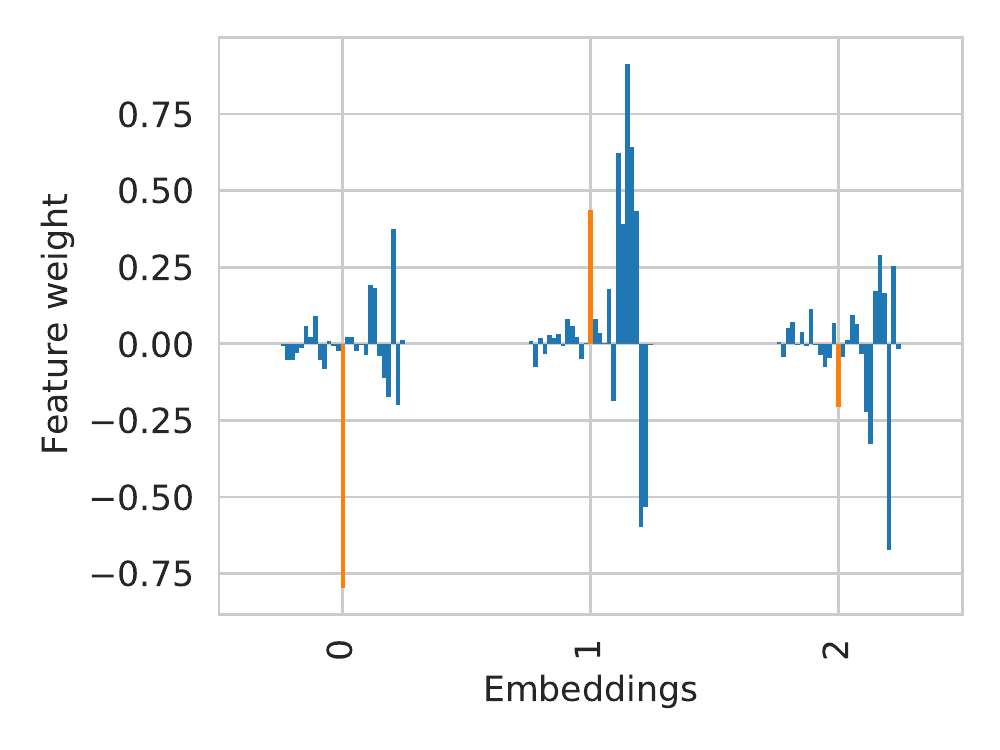}
    \caption{Embedding weights for a linear projection of the Ant state space down to $\mathbb{R}^3$. Only weights for the first 27 (standardized) state features are visualized since the remaining 84 are always zero. The feature that corresponds to (whole body) velocity in the same direction as the main environment objective is highlighted in orange.}
    \label{fig:ant-enc3-features}
\end{figure}

To better understand the results of the Ant experiment, we investigate the projection matrix learned at the start of the algorithm. Figure \ref{fig:ant-enc3-features} gives a representative example of a projection learned for an ENC(3) run.
In the diagram, each bar indicates the impact each feature of the state space has on the final embedding.
The orange bar highlights the feature corresponding to linear torso velocity in the x-direction, i.e. the direction in which the extrinsic objective rewards an agent for running in.
All the bars to the left correspond to joint configurations, link orientations, and all the bars to the right correspond to other velocities.

The feature for velocity in the target direction is well represented. 
However, so are the features for the 8 joint velocities (8 rightmost bars in each group).
Since it is a lot easier to move a single joint than to coordinate all of them for locomotion, the algorithm might more easily converge to this strategy than figure out a way to walk.
Moreover, because the projection mixes features for movement of single joints with features for locomotion of the entire body, it becomes more difficult for the classifier to distinguish the two. For instance, an ant that figures out how to walk may (in the projected space) look similar to one that only twitches some of its joints.

%%%%%%%%%%%%%%%%%%%%%%%%%%%%%%%%%%%%%%%%%%%%%%%%%%%%%%%%%%%%%%%%%%%%%%%%%%%%%%%%%%%%%%%
\section{Conclusion}
\label{sec:conclusion}
In this work, we propose a data-driven approach for guiding skill discovery towards learning useful behaviors in complex and high-dimensional spaces. 
Using examples of expert data, we fit a state-space projection that preserves information that makes expert behavior recognizable.
The projection helps discover better behaviors by ensuring that skills similar to the expert are distinguishable from randomly initialized skills. 
We show the applicability of our approach in a variety of RL tasks, ranging from a simple 2D point maze problem to continuous control locomotion. 
For future work, we aim to improve the embedding scheme of the state projection to be suitable for a wider range of environments.

\paragraph{Acknowledgment.} We would like to thank Kerstin Bach and Rudolf Mester for their useful feedback.

\bibliographystyle{icml2021}
\bibliography{bibliography.bib}

\clearpage
\appendix

%%%%%%%%%%%%%%%%%%%%%%%%%%%%%%%%%%%%%%%%%%%%%%%%%%%%%%%%%%%%%%%%%%%%%%%%%%%%%%%%%%%%%%%
\section{Pseudocode}
\begin{algorithm}[H]
  \scriptsize
\DontPrintSemicolon
\SetAlgoLined
\SetKwInOut{Input}{input}
\SetKwInOut{Output}{output}
\Input{Replay buffer $\mathcal{B}$, dataset of expert states $\mathcal{D}$. }

Initialize policy ${\pi}_{\vartheta}$, Q-functions $\{Q^1_{\theta}, Q^2_{\theta}\}$,  discriminator $q_{\phi}$, encoder $f_{\chi}$. \;

\SetKwFunction{FMain}{Pre-train-Encoder}
\SetKwProg{Fn}{Function}{:}{}
\Fn{\FMain{$f_{\chi}, \mathcal{D}$}}{
    Initialize classifier $h_{\psi}$.\;
    
    Sample actions from a random policy: $a_t \sim \pi_{rand}(a_t|s_t)$. \;
    
    Step environment using the random actions: $s_{t+1} \sim  p(s_{t+1}|s_t, a_t)$.\;
    
    Add visited states to the dataset: $\mathcal{D} \leftarrow  \mathcal{D} \cup \left\{(s_{t+1}, 0) \right\}$.\;
    
    Update encoder ($f_\chi$) through joint training with the classifier $h_{\psi}$ to maximize the likelihood of $\mathcal{D}$. \;
    
    Discard the classifier $h_{\psi}$. \;
    
    \KwRet $f_{\chi}$\; }
\For{$epoch \leftarrow 1$ \KwTo $num\_of\_epochs$}{
    \For{$ t \leftarrow 1$ \KwTo $environment\_steps\_per\_epoch$}{
        Sample a skill: $ z \sim p(z) $. \;
        
        Sample action from the skill-conditioned policy: $a_t \sim \pi_{\vartheta}(a_t|s_t, z)$.\;
        
        Step environment: $s_{t+1} \sim  p(s_{t+1}|s_t, a_t)$.\;
        
        Add a transition to the replay buffer: $\mathcal{B} \leftarrow  \mathcal{B} \cup \left\{(s_t, z, a_t, s_{t+1}) \right\}$.\;
    }
    \For{$i \leftarrow 1$ \KwTo $train\_steps\_per\_epoch$}{
        Sample batch of transitions from the replay buffer: $M \sim \mathcal{B}$.\;
        
        Encode states: $e_{t+1} = f_{\chi} (s_{t+1})$. \;
        
        Compute intrinsic reward: $ r = \log q_{\phi}(z|e_{t+1}) - \log p(z) $. \;
        
        Update Q-functions $\{Q^1_{\theta}, Q^2_{\theta}\}$ to minimize the soft Bellman residual.\;
        
        Update policy $ \pi_{\vartheta}$ using the minimum of the two Q-functions. \;
        
        Update discriminator $q_{\phi}$ with MLE. \;
    }
}
\caption{Skill Discovery with Data-Driven Guidance}
\Output{ Learned skill-conditioned policy $\pi_{\vartheta}$.}
\label{alg:alg}
\end{algorithm}

%%%%%%%%%%%%%%%%%%%%%%%%%%%%%%%%%%%%%%%%%%%%%%%%%%%%%%%%%%%%%%%%%%%%%%%%%%%%%%%%%%%%%%%
\section{Additional Experimental Details}

\begin{figure*}[!htb]
    \centering
    \includegraphics[trim=0 20 0 10, width=\textwidth]{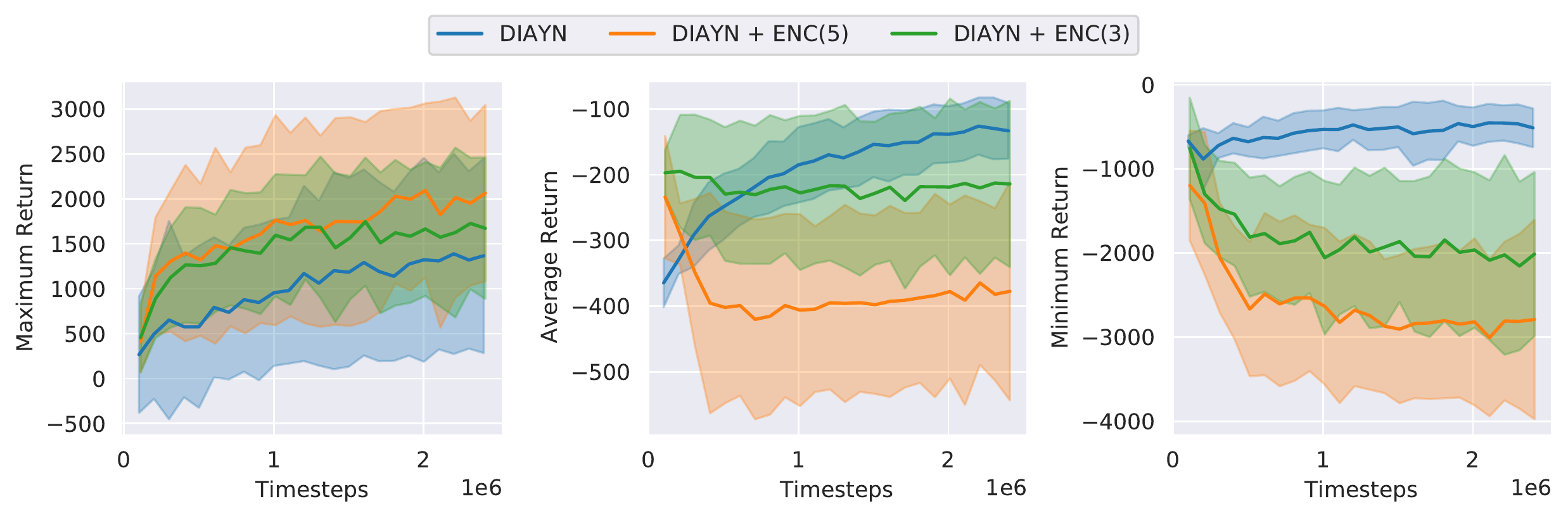}
    \caption{Maximum, average, and minimum return (computed over all 50 skills) for \textbf{HalfCheetah-v2} during training. Shaded areas correspond to $\pm$ standard deviation across 5 random seeds.}
    \label{fig:train-ret-cheetah}
\end{figure*}

\begin{figure*}[!htb]
    \centering
    \includegraphics[trim=0 20 0 10, width=\textwidth]{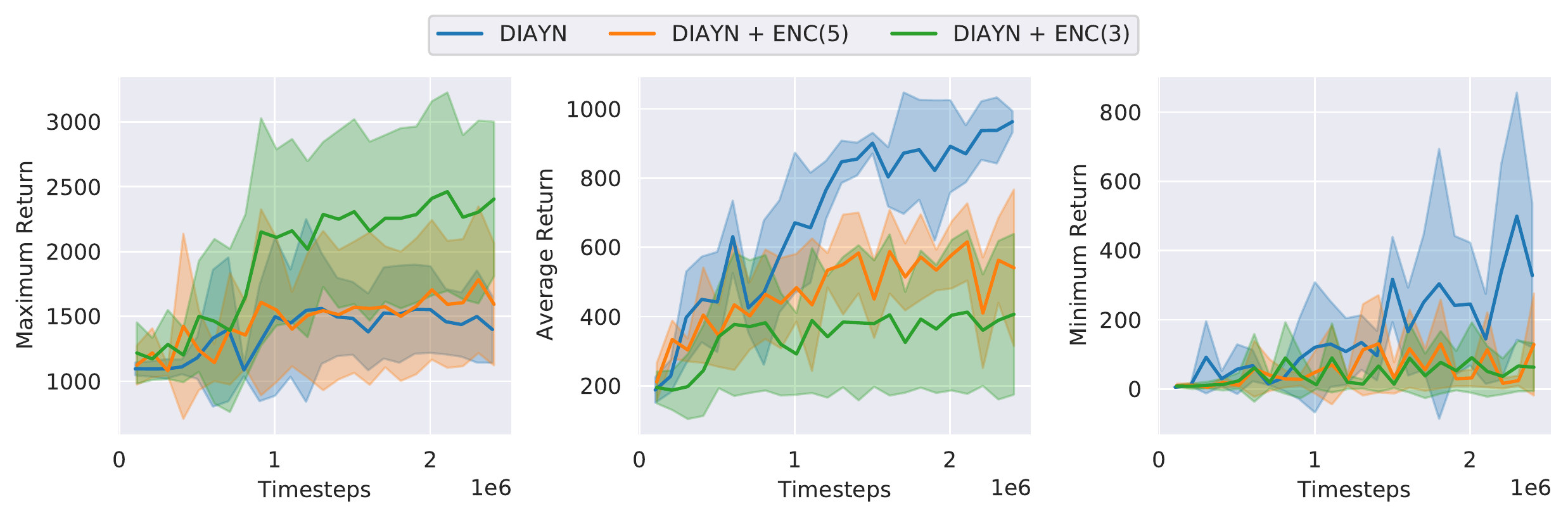}
    \caption{Maximum, average, and minimum return (computed over all 50 skills) for \textbf{Hopper-v2} during training. Shaded areas correspond to $\pm$ standard deviation across 5 random seeds.}
    \label{fig:train-ret-hopper}
\end{figure*}

\begin{figure*}[!htb]
    \centering
    \includegraphics[trim=0 20 0 10, width=\textwidth]{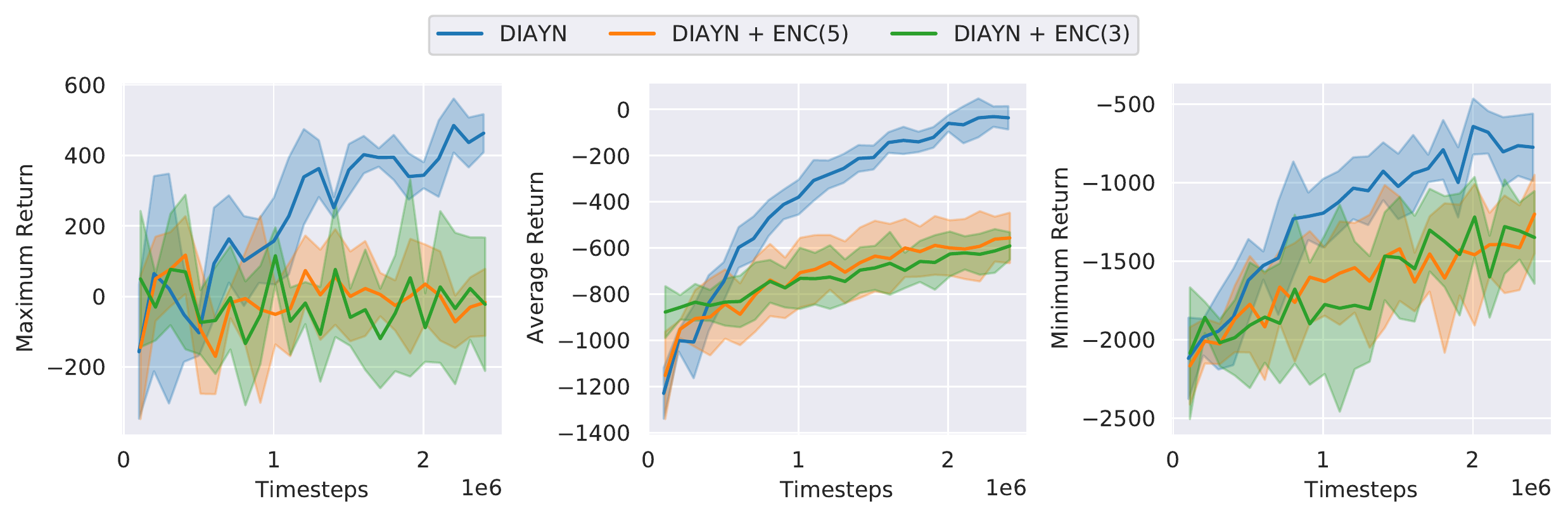}
    \caption{Maximum, average, and minimum return (computed over all 50 skills) for \textbf{Ant-v2} during training. Shaded areas correspond to $\pm$ standard deviation across 5 random seeds.}
    \label{fig:train-ret-ant}
\end{figure*}

This appendix extends \ref{sec:experiments} with additional plots and commentary.
\autoref{fig:train-ret-cheetah}, \ref{fig:train-ret-hopper}, \ref{fig:train-ret-ant} show maximum, average and minimum return for the three environments.

%%%%%%%%%%%%%%%%%%%%%%%%%%%%%%%%%%%%%%%%%%%%%%%%%%%%%%%%%%%%%%%%%%%%%%%%%%%%%%%%%%%%%%%
\section{Implementation Details}

Conceptually, our skill-discovery algorithm is the same as DIAYN \cite{eysenbach2018diversity}.
There are, however, a few implementation differences that we empirically found to work just as well. 
Below follows a brief rundown of the key implementation details of the algorithm used in the documented experiments.

\begin{enumerate}
    \item Two Q-functions $Q^1_\theta(s,a)$ \& $Q^2_\theta(s,a)$ are used, both with target clones $Q^1_{\theta'}$ \& $Q^2_{\theta'}$ that are continuously updated with polyak averaging. Both Q-functions attempt to predict the same quantity:
    \begin{align*}
        Q^{1,2}_\theta(s_t, a_t) &= \mathbb{E}_{s,a \sim \pi_\theta}[r(s_t, a_t) \\
        &+ \sum_{t' > t} \gamma^{t'-t}(\alpha \mathcal{H}(a_{t'}) + r(s_{t'}, a_{t'}))]
    \end{align*}
    
    \item The policy distribution is a mixture of Gaussians with four components. The policy network predicts the mixture logits, as well as the means and log standard deviations of the Gaussians. The output is squashed through a hyperbolic tangent function, similar to \cite{haarnoja2018soft}.
    
    \item The policy is updated by climbing the gradient of the minimum of the two Q functions (DDPG-style  (Lilli-crap et al.)).
    \begin{align*}
        J(\pi_\theta) = \min_{i \in \{1, 2\}} Q^i_{\theta}(\pi_\theta(a|s)) + \alpha \mathcal{H}(\pi_\theta(a|s))
    \end{align*}
    This requires that the actions sampled from the policy are differentiable. Each gaussian component of the mixture is reparametrized the standard way, and the mixture is reparametrized with Gumbel-Softmax \cite{jang2017categorical}.
    
    \item $Q^{1,2}_\theta$ is trained by descending on the squared temporal difference (TD) errors generated by the minimum of the target networks $Q^1_{\theta'}$ \& $Q^2_{\theta'}$
    \begin{align*}
        TD(s, a, r, s') &= Q^{1,2}_\theta(s, a) - r - \gamma (\\&\min_{i \in \{1,2\}} Q^i_{\theta'}(s', \pi_\theta(a'|s')) + \alpha \mathcal{H}(\pi_\theta(a'|s'))\\&)
    \end{align*}
    
\end{enumerate}

\end{document}